\documentclass{article}

 \usepackage[dblblindworkshop, final]{neurips_2025}

\usepackage[utf8]{inputenc} % allow utf-8 input
\usepackage[T1]{fontenc}    % use 8-bit T1 fonts
\usepackage{hyperref}       % hyperlinks
\usepackage{url}            % simple URL typesetting
\usepackage{booktabs}       % professional-quality tables
\usepackage{amsfonts}       % blackboard math symbols
\usepackage{nicefrac}       % compact symbols for 1/2, etc.
\usepackage{microtype}      % microtypography
\usepackage{xcolor}         % colors
\usepackage{multirow}
\usepackage{array}
\usepackage{float}
\usepackage{longtable}
\usepackage{array}
\usepackage{longtable}
\usepackage{graphicx}
\usepackage{subcaption}
\usepackage[most]{tcolorbox}
\usepackage{caption}
\usepackage{textcomp}

\usepackage{tikz}
\usepackage{fontawesome5}
\usetikzlibrary{arrows.meta,positioning,shapes.misc,shapes.symbols,shapes.geometric,fit,backgrounds,calc}
\usepackage{adjustbox}

\title{Adapting General-Purpose Foundation Models for X-ray Ptychography in Low-Data Regimes}

\author{%
  Robinson Umeike\thanks{Work performed while at Argonne National Laboratory.} \\
  The University of Alabama\\
  \texttt{crumeike@crimson.ua.edu} \\
  \And
  Neil Getty \\
  Argonne National Laboratory \\
  \texttt{ngetty@anl.gov} \\
  \AND
  Yin Xiangyu \\
  Argonne National Laboratory \\
  \texttt{xyin@anl.gov} \\
  \And
  Yi Jiang \\
  Argonne National Laboratory \\
  \texttt{yjiang@anl.gov} \\
}

\begin{document}

\maketitle

\begin{abstract}
    The automation of workflows in advanced microscopy is a key goal where foundation models like Language Models (LLMs) and Vision-Language Models (VLMs) show great potential. However, adapting these general-purpose models for specialized scientific tasks is critical, and the optimal domain adaptation strategy is often unclear. To address this, we introduce PtychoBench, a new multi-modal, multi-task benchmark for ptychographic analysis. Using this benchmark, we systematically compare two specialization strategies: Supervised Fine-Tuning (SFT) and In-Context Learning (ICL). We evaluate these strategies on a visual artifact detection task with VLMs and a textual parameter recommendation task with LLMs in a data-scarce regime. Our findings reveal that the optimal specialization pathway is task-dependent. For the visual task, SFT and ICL are highly complementary, with a fine-tuned model guided by context-aware examples achieving the highest mean performance (Micro-F1 of 0.728). Conversely, for the textual task, ICL on a large base model is the superior strategy, reaching a peak Micro-F1 of 0.847 and outperforming a powerful "super-expert" SFT model (0-shot Micro-F1 of 0.839). We also confirm the superiority of context-aware prompting and identify a consistent \textit{contextual interference} phenomenon in fine-tuned models. These results, benchmarked against strong baselines including GPT-4o and a DINOv3-based classifier, offer key observations for AI in science: the optimal specialization path in our benchmark is dependent on the task modality, offering a clear framework for developing more effective science-based agentic systems.
\end{abstract}

\section*{Introduction}
Ptychography is a popular characterization technique used across materials science [1,2], from semiconductors to biological specimens [3-6]. It computationally leverages redundant information in measured data to reconstruct the sample's structure at a spatial resolution far surpassing the limits of physical lenses. However, the quality of a ptychographic reconstruction depends on a complex, multi-dimensional optimization of experimental and algorithmic parameters. This process has typically relied on expert intuition and time-consuming, trial-and-error workflows. Recent work [7], demonstrated that agentic workflows orchestrating multiple Large Language Model (LLMs) and Vision-Language Model (VLM) can facilitate more streamlined and automated data analysis. The workflow includes a central Ptychography Agent (LLM) that gathers experimental context and recommends parameters, a Coding Agent that generates and execute reconstruction scripts, and a Diagnosis Agent (VLM) that visually assesses the reconstructed images for quality issues.

Although these agentic frameworks have demonstrated the viability of an LLM-driven workflow, the performance of their two core decision-making components, the VLM-based Diagnosis Agent and the LLM-based Ptychography Agent, had not been rigorously benchmarked. Their effectiveness, especially with new types of structures or artifacts that constitute out-of-distribution data for general-purpose models, remained a critical open question. To address this gap and systematically evaluate these agents, this paper makes two key contributions. First, we introduce PtychoBench, a novel, expert-annotated multi-modal, multi-task benchmark dataset for X-ray ptychographic analysis. This benchmark is designed to independently evaluate the two key agents with tasks of artifact detection for the VLM and parameter recommendation for the LLM, which is annotated on a smaller, more complex subset of the data.

This benchmark enables a systematic investigation into two distinct approaches for specializing these models: parameter-based specialization via supervised fine-tuning (SFT) [8-10] and in-context learning (ICL) [11-13]. We evaluate these strategies using open-weight Llama models of various scales (8B to 90B) and their fine-tuned counterparts [14], benchmarking their performance against a leading proprietary model like GPT-4o [15] and traditional state-of-the-art vision classifiers using DINOv3 embeddings [16]. This allows us to dissect the relationship between these two learning paradigms across different modalities and task complexities.

Our findings reveal two distinct, task-dependent pathways to expertise within our data-scarce setting. For the VLM-based artifact detection task, we find SFT and ICL are highly complementary, while for the LLM-based parameter tuning task, ICL on a large base model is the superior strategy, outperforming an overspecialized fine-tuned expert. Across all experiments, we identify a consistent \textit{contextual interference} phenomenon where irrelevant context degrades SFT model performance. These observations highlight a critical trade-off between training-time and inference-time specialization, underscoring that the relevance of retrieved data is a primary driver of success. This work provides both a benchmark and a clear direction for future research, motivating the application of advanced algorithms like Group Relative Policy Optimization (GRPO) to enhance model reasoning [17].

\begin{figure}[h!]
\centering
\begin{tikzpicture}[scale=0.7, transform shape,
    node distance=0.8cm and 1.1cm,
    box/.style={rectangle, rounded corners, minimum width=2.2cm, minimum height=0.5cm, text centered, draw=black, fill=blue! 10, font=\small},
    databox/.style={rectangle, rounded corners, minimum width=2.2cm, minimum height=0.7cm, text centered, draw=black, fill=green! 10, font=\small},
    taskbox/.style={rectangle, rounded corners, minimum width=2.2cm, minimum height=0.7cm, text centered, draw=black, fill=orange! 15, font=\small},
    modelbox/.style={rectangle, rounded corners, minimum width=2cm, minimum height=0.6cm, text centered, draw=black, fill=purple! 10, font=\scriptsize},
    arrow/.style={-Stealth, thick},
    dashedarrow/.style={-Stealth, thick, dashed},
    label/.style={font=\scriptsize\bfseries} 
]

% Data Source
\node[databox] (data) {\faDatabase\ \textbf{394 Samples} \par [APS Experiments]};

% Data Processing
\node[databox, below=0.6cm of data] (clean) {\textbf{391 Samples} \par [Cleaned Dataset]};
\node[databox, below left=0.6cm and 0.5cm of clean] (vlmdata) {\textbf{Task 1: VLM} (312 train / 79 test)};
\node[databox, below right=0.6cm and 0.5cm of clean] (llmdata) {\textbf{Task 2: LLM} (91 train / 44 test)};

% Task Definitions
\node[taskbox, below=0.4cm of vlmdata, xshift=1.0cm] (task1) {\faImage\ \textbf{Artifact Detection }  (Multi-modal)};
\node[taskbox, below=0.4cm of llmdata, xshift=-1.0cm] (task2) {\faChartBar\ \textbf{Parameter Recommendation } (Text-only)};

% Specialization Strategies
\node[box, below=0.8cm of task1, xshift=-1.6cm] (sft1) {\faCog\ \textbf{SFT: }\par LoRA (r=16)};
\node[box, below=0.8cm of task1, xshift=1.6cm] (icl1) {\faLightbulb\ \textbf{ICL: }(RFS / SSFS)};

\node[box, below=0.8cm of task2, xshift=-1.6cm] (sft2) {\faCog\ \textbf{SFT: }LoRA (r=16)};
\node[box, below=0.8cm of task2, xshift=1.6cm] (icl2) {\faLightbulb\ \textbf{ICL: }(RFS / SSFS)};

% Models
\node[modelbox, below=0.7cm of sft1] (vlm11) {Llama3.2-Vision \par 11B/90B};
\node[modelbox, below=0.7cm of icl1] (vlmbase) {Base + Context \par (k=\{0,1,3,5,7\})};

\node[modelbox, below=0.7cm of sft2] (llm8) {Llama3.1-Instruct \par 8B/70B};
\node[modelbox, below=0.7cm of icl2] (llmbase) {Base + Context (k=\{0,1,3,5,7\})};

% Baselines
\node[modelbox, below=8.3cm of data, xshift=-1.6cm] (gpt4o) {\textcolor{blue}{\faRobot\ GPT-4o}};
\node[modelbox, below=8.3cm of data, xshift=1.6cm] (dino) {\textcolor{purple}{DINOv3}};

% Evaluation
\node[box, below=7.2cm of data, fill=red!15] (eval) {\faChartBar\ \textbf{Evaluation} \par (Micro-F1)};

% Arrows - Data Flow
\draw[arrow] (data) -- (clean);
\draw[arrow] (clean) -| (vlmdata);
\draw[arrow] (clean) -| (llmdata);
\draw[arrow] (vlmdata) -- (task1);
\draw[arrow] (llmdata) -- (task2);

% Arrows - Specialization
\draw[arrow] (task1) -- (sft1);
\draw[arrow] (task1) -- (icl1);
\draw[arrow] (task2) -- (sft2);
\draw[arrow] (task2) -- (icl2);

% Arrows - Models
\draw[arrow] (sft1) -- (vlm11);
\draw[arrow] (icl1) -- (vlmbase);
\draw[arrow] (sft2) -- (llm8);
\draw[arrow] (icl2) -- (llmbase);

% Arrows - Evaluation
\draw[arrow] (vlm11) |- (eval);
\draw[arrow] (vlmbase) |- (eval);
\draw[arrow] (llm8) |- (eval);
\draw[arrow] (llmbase) |- (eval);
\draw[dashedarrow] (gpt4o) -- (eval);
\draw[dashedarrow] (dino) -- (eval);

% Labels
\node[label, above=0.1cm of data] {Data Curation};
\node[label, left=0.1cm of vlmdata, align=right] {80/20\\ Split};
\node[label, right=0.1cm of llmdata, align=left] {Filtered\\ Subset};
\node[label, left=0.3cm of sft1, rotate=90, anchor=south] {Specialization};
\node[label, right=0.0001cm of gpt4o, align=center] {Baselines};

% Grouping boxes
\begin{scope}[on background layer]
    \node[draw=blue!50, thick, dashed, rounded corners, fit={(task1) (sft1) (icl1) (vlm11) (vlmbase)}, inner sep=0.2cm, fill=blue!3] {};
    \node[draw=orange!50, thick, dashed, rounded corners, fit={(task2) (sft2) (icl2) (llm8) (llmbase)}, inner sep=0.2cm, fill=orange!3] {};
\end{scope}

\end{tikzpicture}
\caption{\textbf{Methodology Overview}: PtychoBench workflow from data curation through specialization strategies (SFT and ICL) to evaluation. The dataset is partitioned for two tasks: VLM-based artifact detection (blue) and LLM-based parameter recommendation (orange). Both tasks are evaluated using Micro-F1 against baselines including GPT-4o and DINOv3 (see supplementary material for more details).}
\label{fig:methodology}
\end{figure}
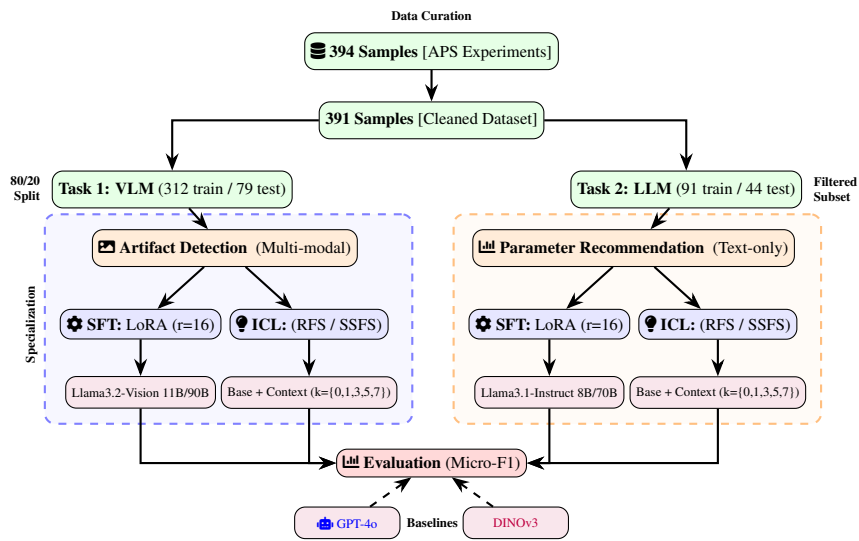

\vspace{-1.6em}

\section{Methodology and Experimental Design}
\label{headings}
This study, illustrated in Figure~\ref{fig:methodology}, systematically evaluates two specialization strategies using the PtychoBench dataset. The benchmark contains 391 expert-annotated samples for a multi-modal Artifact Detection (VQA) task and a 135-sample subset for a text-only Parameter Recommendation (QA) task. The full dataset was first partitioned into an 80/20 train/test split, from which the task-specific subsets were then derived.

We investigate two primary specialization strategies. The first is Supervised Fine-Tuning (SFT) [23], which finds specialized parameters $\theta_{SFT}$ by minimizing a loss function $\mathcal{L}$ over the training data $D_{train}$. For this, we employed a parameter-efficient approach using Low-Rank Adaptation (LoRA) with a rank $(r)$ of 16 [21] and the AdamW 8-bit optimizer [24]. The second strategy is in-Context Learning (ICL), where a model's prediction $\hat{Y}$ for a test input $x_{test}$ is conditioned on a context set of $k$ examples, $C=\{(x_i, y_i)\}_{i=1}^k$, such that $\hat{Y}=f_\theta(x_{test} | C)$ [11]. We tested two context selection protocols for $k=\{0, 1, 3, 5, 7\}$: the Random Few-Shot (RFS), where examples are sampled randomly, and Sample-Specific Few-Shot (SSFS), where examples are selected based on matching sample type data. 

 This comprehensive setup allows us to investigate key questions relevant to deploying specialized AI agents for X-ray ptychography: \textbf{(1)} The optimal specialization strategy (SFT vs. ICL) for robust deployment in a data-scarce setting; \textbf{(2)} The impact of contextual relevance (SSFS vs. RFS) on agent performance; \textbf{(3)}  Whether the distinct systems in an autonomous workflow require different specialization strategies; and \textbf{(4)} The effect of model scale on performance.

To investigate these questions, our evaluation includes open-weight Llama models at multiple scales (11B/90B for VLM, 8B/70B for LLM) and two powerful external benchmarks, GPT-4o and a DINOv3-based classifier. We also established two crucial internal baselines to contextualize performance: 1) 0-Shot Performance to measure the standalone capability of each model; and 2) the RFS strategy to provide a naive ICL baseline that quantifies the value of our retrieval strategy. The primary metric for all experiments is the Micro-F1 score. Complete hyperparameter configurations, prompt templates, and a detailed dataset breakdown are provided in the supplementary material.

\vspace{-1.4em}

\begin{table}[h!]
\centering
\tiny % Reduces the font size of the table
\caption{VLM Performance on Artifact Detection (Mean $\pm$ Std)} 
\label{table:artifact_detection_mean_std}
\setlength{\arrayrulewidth}{0.7pt} % Sets the width of all horizontal and vertical rules
\setlength{\tabcolsep}{7pt} % Adds horizontal space between columns
\renewcommand{\arraystretch}{1.2} % Adds vertical space between rows
\begin{tabular}{|l|c|c|c|c|c|c|}
\hline
\textbf{Model} & \textbf{Fewshot Strategy} & \textbf{0-shot} & \textbf{1-shot} & \textbf{3-shot} & \textbf{5-shot} & \textbf{7-shot} \\
\hline
\multirow{2}{*}{\textbf{SFT Llama 3.2-Vision 11B}} & RFS & \multirow{2}{*}{0.493 $\pm$ 0.041} & 0.322 $\pm$ 0.036 \textcolor{red}{$\downarrow$} & 0.319 $\pm$ 0.041 & 0.389 $\pm$ 0.041 & 0.395 $\pm$ 0.041 \\
\cline{2-2}\cline{4-7}
& SSFS & & \textbf{0.547 $\pm$ 0.044} & \textbf{0.636 $\pm$ 0.044} & \textbf{0.667 $\pm$ 0.043} & \textbf{0.713 $\pm$ 0.038} \\
\hline\hline
\multirow{2}{*}{\textbf{Base Llama 3.2-Vision 11B}} & RFS & \multirow{2}{*}{0.219 $\pm$ 0.025} & 0.254 $\pm$ 0.034 & 0.250 $\pm$ 0.035 & 0.313 $\pm$ 0.033 & 0.342 $\pm$ 0.036 \\
\cline{2-2}\cline{4-7}
& SSFS & & \textbf{0.609 $\pm$ 0.047} & \textbf{0.628 $\pm$ 0.046} & \textbf{0.696 $\pm$ 0.041} & \textbf{0.694 $\pm$ 0.039} \\
\hline\hline
\multirow{2}{*}{\textbf{SFT Llama 3.2-Vision 90B}} & RFS & \multirow{2}{*}{0.430 $\pm$ 0.041} & 0.333 $\pm$ 0.036 \textcolor{red}{$\downarrow$} & 0.409 $\pm$ 0.036 & 0.401 $\pm$ 0.037 & 0.446 $\pm$ 0.035 \\
\cline{2-2}\cline{4-7}
& SSFS & & \textbf{0.586 $\pm$ 0.043} & \textbf{0.596 $\pm$ 0.046} & \textbf{0.671 $\pm$ 0.041} & \textbf{0.728 $\pm$ 0.036} \textcolor{green}{$^{\star}$}\\
\hline\hline
\multirow{2}{*}{\textbf{Base Llama 3.2-Vision 90B}} & RFS & \multirow{2}{*}{0.019 $\pm$ 0.014} & 0.255 $\pm$ 0.034 & 0.176 $\pm$ 0.035 & 0.090 $\pm$ 0.028 & 0.151 $\pm$ 0.038 \\
\cline{2-2}\cline{4-7}
& SSFS & & \textbf{0.610 $\pm$ 0.048} & \textbf{0.628 $\pm$ 0.050} & \textbf{0.623 $\pm$ 0.049} & \textbf{0.706 $\pm$ 0.039} \\
\hline\hline
\multirow{2}{*}{\textbf{\textcolor{blue}{GPT-4o}}} & RFS & \multirow{2}{*}{0.175 $\pm$ 0.026} & 0.163 $\pm$ 0.023 & 0.189 $\pm$ 0.027  & 0.248 $\pm$ 0.030 & 0.280 $\pm$ 0.039 \\
\cline{2-2}\cline{4-7}
& SSFS & & \textbf{0.334 $\pm$ 0.037} & \textbf{0.529 $\pm$ 0.045} & \textbf{0.635 $\pm$ 0.044} & \textbf{0.656 $\pm$ 0.043} \\
\hline\hline
\textbf{\textcolor{purple}{DINOv3}} & - & 0.628 & - & - & - & - \\
\hline
\end{tabular}
\end{table}

\vspace{-2em}

\begin{table}[h!]
\centering
\tiny % Reduces the font size of the table
\caption{LLM Performance on Parameter Recommendation (Mean $\pm$ Std)}
\label{table:param_reco_mean_std}
\setlength{\arrayrulewidth}{0.8pt} % Sets the width of all horizontal and vertical rules
\setlength{\tabcolsep}{8.3pt} % Adds horizontal space between columns
\renewcommand{\arraystretch}{1.2} % Adds vertical space between rows
\begin{tabular}{|l|c|c|c|c|c|c|}
\hline
\textbf{Model} & \textbf{Fewshot Strategy} & \textbf{0-shot} & \textbf{1-shot} & \textbf{3-shot} & \textbf{5-shot} & \textbf{7-shot} \\
\hline
\multirow{2}{*}{\textbf{SFT Llama 3.1 8B}} & RFS & \multirow{2}{*}{0.470 $\pm$ 0.051} & 0.316 $\pm$ 0.037 \textcolor{red}{$\downarrow$} & 0.416 $\pm$ 0.044 & 0.497 $\pm$ 0.053 & 0.552 $\pm$ 0.039 \\
\cline{2-2}\cline{4-7}
& SSFS & & \textbf{0.318 $\pm$ 0.050} & \textbf{0.602 $\pm$ 0.062} & \textbf{0.661 $\pm$ 0.056} & \textbf{0.697 $\pm$ 0.054} \\
\hline\hline
\multirow{2}{*}{\textbf{Base Llama 3.1 8B}} & RFS & \multirow{2}{*}{0.092 $\pm$ 0.022} & 0.183 $\pm$ 0.037 & 0.271 $\pm$ 0.034 & 0.452 $\pm$ 0.048 & 0.388 $\pm$ 0.047 \\
\cline{2-2}\cline{4-7}
& SSFS & & \textbf{0.411 $\pm$ 0.037} & \textbf{0.602 $\pm$ 0.058} & \textbf{0.667 $\pm$ 0.066} & \textbf{0.709 $\pm$ 0.056} \\
\hline\hline
\multirow{2}{*}{\textbf{SFT Llama 3.1 70B}} & RFS & \multirow{2}{*}{0.839 $\pm$ 0.050} & 0.457 $\pm$ 0.047 \textcolor{red}{$\downarrow$} & 0.493 $\pm$ 0.041 & 0.527 $\pm$ 0.049 & 0.595 $\pm$ 0.049 \\
\cline{2-2}\cline{4-7}
& SSFS & & \textbf{0.518 $\pm$ 0.042}  & \textbf{0.608 $\pm$ 0.050} & \textbf{0.683 $\pm$ 0.044} & \textbf{0.695 $\pm$ 0.047} \\
\hline\hline
\multirow{2}{*}{\textbf{Base Llama 3.1 70B}} & RFS & \multirow{2}{*}{0.228 $\pm$ 0.020} & 0.267 $\pm$ 0.038 & 0.362 $\pm$ 0.045 & 0.546 $\pm$ 0.055 & 0.575 $\pm$ 0.056 \\
\cline{2-2}\cline{4-7}
& SSFS & & \textbf{0.543 $\pm$ 0.048} & \textbf{0.746 $\pm$ 0.040} & \textbf{0.784 $\pm$ 0.042} & \textbf{0.847 $\pm$ 0.042} \textcolor{green}{$^{\star}$} \\
\hline\hline
\multirow{2}{*}{\textbf{\textcolor{blue}{GPT-4o}}} & RFS & \multirow{2}{*}{0.313 $\pm$ 0.030} & 0.340 $\pm$ 0.038 & 0.324 $\pm$ 0.044  & 0.518 $\pm$ 0.048 & 0.555 $\pm$ 0.050 \\
\cline{2-2}\cline{4-7}
& SSFS & & \textbf{0.617 $\pm$ 0.048} & \textbf{0.709 $\pm$ 0.047} & \textbf{0.766 $\pm$ 0.046} & \textbf{0.781 $\pm$ 0.039} \\
\hline
\end{tabular}
\end{table}

\vspace{-1.1em}

\section{Results and Analysis}
Our experiments reveal a set of consistent findings regarding the importance of contextual relevance, alongside a striking divergence in optimal specialization strategies between the visual and textual tasks. The full performance metrics for Artifact Detection and Parameter Recommendation are presented in Table 1 and Table 2, respectively, with a comprehensive breakdown of all metrics, including 95\% confidence intervals computed via bootstrapping ($n=10,000$), provided in the Supplementary Material.
\paragraph{A. The Superiority of Context Relevant Prompting.}
The most consistent finding across all models and tasks is the dramatic superiority of Sample-Specific Few-Shot (SSFS) prompting over Random Few-Shot (RFS). Providing contextually relevant examples consistently yields a significant performance boost. For instance, in the artifact detection task (Table 1), the Base Llama 3.2-Vision 90B model’s Micro-F1 score at 3-shots leaps from 0.176 $\pm$ 0.035 with RFS to 0.628 $\pm$ 0.050 with SSFS, a relative increase of 256\%. This power of sample-specific context is most pronounced in base models, where a single relevant example can transform a model from non-functional to highly effective; the Base Llama 3.2-Vision 90B model’s performance jumps from a 0-shot F1-score of 0.019 $\pm$ 0.014 to 0.610 $\pm$ 0.048 after seeing just one relevant sample. Conversely, the results highlight the danger of irrelevant context, indicated by performance drops in Tables 1 and 2 (\textcolor{red}{↓}). For fine-tuned models, RFS often leads to \textit{contextual interference}, where performance degrades below the 0-shot baseline. This is evident in the SFT Llama 3.2-Vision 11B model, where a single random example causes the F1 score to drop from 0.493 $\pm$ 0.041 to 0.322 $\pm$ 0.036, a relative performance decrease of 35\%. This counter-intuitive \textit{contextual interference} phenomenon suggests that a single, irrelevant example can create ambiguity that overrides the model's specialized SFT training, making it a critical failure case to consider.

\paragraph{B. Task-Dependent Specialization Strategies.}
Aligning with established findings that context relevance is critical for zero- or few-shot performance, as demonstrated by the success of techniques such as Retrieval-Augmented Generation (RAG) [9, 12, 13, 20, 22], our results show that the optimal strategy to achieving peak performance is largely dependent on the task's modality.

\textbf{I. Visual Material Diagnosis.} For the artifact detection task, the highest mean performance is achieved when SFT and ICL are used together, with the SFT Llama 3.2-Vision 90B model and 7-shot SSFS reaching a peak Micro-F1 of 0.728 $\pm$ 0.036 (Table 1). This outperforms the best-performing base model (Base 90B with 7-shot SSFS at 0.706 $\pm$ 0.039), demonstrating a clear combined effect where fine-tuning primes the model to better leverage in-context examples. The value of this domain-specific priming is highlighted by a comparison with the generalist model, GPT-4o. While GPT-4o is a capable in-context learner (reaching 0.656 $\pm$ 0.043 with SSFS), its 0-shot performance is a low 0.175 $\pm$ 0.026, far below that of our specialized SFT models (e.g., SFT-11B at 0.493 $\pm$ 0.041). This gap underscores the necessity of domain adaptation for this task. Ultimately, the peak score of SFT-90B + SSFS strategy surpasses that of the guided GPT-4o, confirming that for complex visual analysis, a specialized model combined with relevant context remains the state-of-the-art approach. However, SFT in a low-data regime reveals a paradox of scale: the smaller SFT-11B model has a stronger 0-shot performance than the larger SFT-90B (0.493 $\pm$ 0.041 vs. 0.430 $\pm$ 0.041). We attribute this to our training protocol: the number of training steps for each model was determined by monitoring performance on a validation set to prevent overfitting. The larger 90B model demonstrated faster convergence, and thus was trained for fewer steps ($\approx$ 20 epochs) to reach its optimal performance compared to the 11B model (50 epochs). Despite its lower baseline, the SFT-90B model shows a higher potential that is unlocked by ICL; its performance increases by over 69\% from its 0-shot baseline when provided with 7 specific examples, a much larger gain than that seen for the SFT-11B model(45\%).

\textbf{II. Textual Recommendation.} In stark contrast to the visual task, the results show that the most effective strategy is In-Context Learning on a large, general-purpose base model. The peak performance across all experiments was achieved by the Base Llama 3.1-70B model with 7-shot SSFS, reaching a Micro-F1 of 0.847 $\pm$ 0.042 (Table 2). This strategy not only surpassed all fine-tuned configurations but also outperformed the GPT-4o baseline, which peaked at 0.781 $\pm$ 0.039 under the same conditions. The fine-tuned models exhibit a fascinating and contrary behavior. The SFT-70B model emerges as a specialized "super-expert," achieving an exceptional 0-shot performance of 0.839 $\pm$ 0.050. This indicates that while SFT was highly effective at instilling expert knowledge, this knowledge became rigid; any form of ICL, even with relevant sample-specific examples, consistently degraded its performance. This finding, where more context hurts performance, highlights a key risk of over-specialization in a low-data regime. This antagonistic relationship between SFT and ICL for this task suggests that for certain reasoning pathways, fine-tuning may create a rigid model that is easily disturbed by external context. Finally, the results also underscore the critical role of scale for ICL on base models. The peak performance of the Base-70B model (0.847 $\pm$ 0.042) represents a significant leap over the Base-8B model (0.709 $\pm$ 0.056), confirming that for this text-based reasoning task, a larger model is substantially more capable of leveraging in-context examples to achieve expert-level performance.
\vspace{-0.5em}
\section{Discussion and Conclusion}

A key finding from our work is that the optimal strategy for specializing AI models in our benchmark is not universal but appears to be dependent on the task's modality. Our results show two distinct pathways to expert performance: a complementary relationship between SFT and ICL for the visual-perceptual task, and the supremacy of ICL on a large base model for the textual-reasoning task. We hypothesize this divergence is rooted in what SFT learns in a low-data regime; for visual diagnosis, SFT builds a foundational "visual vocabulary" that ICL can effectively refine, while for the more abstract recommendation task, SFT may create a powerful but inflexible expert by overfitting to specific reasoning paths. We also note that these two tasks differ in data volume and complexity. While our results point to modality as a key factor, further research would be valuable to fully disentangle these variables.

The implications for designing autonomous science systems are significant, suggesting a design guideline where the specialization strategy is tailored to the model’s function. For instance, our VLM results were contextualized against a strong DINOv3-based classifier (micro-F1 of 0.628), which used a 7B model as a feature extractor trained on our full 312-sample training set. The fact that our Base-90B + SSFS strategy (0.706 $\pm$ 0.039), using only 7 examples at inference, achieved a mean performance substantially higher than the fully-trained DINOv3 baseline, highlights the remarkable data efficiency of the ICL paradigm. This demonstrates that a well guided general purpose VLM can effectively match or even exceed the performance of a traditional, fully trained classifier, despite using a fraction of the data. This suggests perceptual AI models may benefit most from a hybrid SFT+ICL approach, while models for downstream reasoning may achieve higher performance and flexibility as large base models guided by a sophisticated retrieval system.

While our findings are robust across multiple model scales, this study is primarily based on the PtychoBench dataset and Llama model family; future work should validate these observations on other scientific domains. Furthermore, our results underscore that high-quality context is paramount. This motivates our future work in exploring data augmentation strategies to improve robustness and applying reinforcement learning algorithms, such as GRPO, to move beyond providing static examples and instead foster genuine, step-by-step reasoning pathways in scientific AI systems.

\newpage
\section{Acknowledgments and Disclosure of Funding}
This work was supported by the Laboratory Directed Research and Development (LDRD) Program at Argonne National Laboratory under Project Number 2025-0495. This research used resources of the Advanced Photon Source, a U.S. Department of Energy (DOE) Office of Science User Facility operated for the DOE Office of Science by Argonne National Laboratory under Contract No. DE-AC02-06CH11357.

\section{Data and Code Availability}
To foster further research in this domain, our code and LoRA adapters are available at \url{https://github.com/crumeike/PtychoBench}. The PtychoBench dataset is available upon request and subject to institutional approval. Requests regarding data access should be directed to \texttt{yjiang@anl.gov}. For questions about the code, please contact \texttt{crumeike@crimson.ua.edu} or \texttt{ngetty@anl.gov}.

% \newpage
\section*{References}

\medskip

\small

\setlength{\parindent}{0pt}
\setlength{\leftskip}{0cm}
\setlength{\parskip}{0.8em}

\noindent\hangindent=0.5cm [1] F. Pfeiffer, (2017) X-ray ptychography. \textit{Nature Photonics} \textbf{12}(1):9--17.

\noindent\hangindent=0.5cm [2] J. Miao, (2025) Computational microscopy with coherent diffractive imaging and ptychography. \textit{Nature} \textbf{637}(8045):281--295.

\noindent\hangindent=0.5cm [3] M. Holler, M. Guizar-Sicairos, E. H. R. Tsai, R. Dinapoli, E. Müller, O. Bunk, J. Raabe, and G. Aeppli, (2017) High-resolution non-destructive three-dimensional imaging of integrated circuits. \textit{Nature} \textbf{543}(7645):402--406.

\noindent\hangindent=0.5cm [4] T. Aidukas, N. W. Phillips, A. Diaz, E. Poghosyan, E. Müller, A. F. J. Levi, G. Aeppli, M. Guizar-Sicairos, and M. Holler, (2024) High-performance 4-nm-resolution X-ray tomography using burst ptychography. \textit{Nature} \textbf{632}(8023):81--88.

\noindent\hangindent=0.5cm [5] Y. Young-Sang, M. Farmand, C. Kim, Y. Liu, C. P. Grey, F. C. Strobridge, T. Tyliszczak, R. Celestre, P. Denes, J. Joseph, H. Krishnan, F. R. N. C. Maia, A. L. D. Kilcoyne, S. Marchesini, T. P. C. Leite, T. Warwick, H. Padmore, J. Cabana, and D. A. Shapiro, (2018) Three-dimensional localization of nanoscale battery reactions using soft X-ray tomography. \textit{Nat Commun} \textbf{9}(1).

\noindent\hangindent=0.5cm [6] J. Deng, Y. H. Lo, M. Gallagher-Jones, S. Chen, A. Pryor Jr, Q. Jin, Y. P. Hong, Y. S. G. Nashed, S. Vogt, J. Miao, and C. Jacobsen, (2018) Correlative 3D x-ray fluorescence and ptychographic tomography of frozen-hydrated green algae. \textit{Sci. Adv.} \textbf{4}(11).

\noindent\hangindent=0.5cm [7] X. Yin, C. Shi, Y. Han, and Y. Jiang, (2024) PEAR: A Robust and Flexible Automation Framework for Ptychography Enabled by Multiple Large Language Model Agents. arXiv.

\noindent\hangindent=0.5cm [8] C. Hyung Won, L. Hou, S. Longpre, B. Zoph, Y. Tay, W. Fedus, E. Li, X. Wang, M. Dehghani, S. Brahma, A. Webson, S. Shane Gu, Z. Dai, M. Suzgun, X. Chen, A. Chowdhery, D. Valter, S. Narang, G. Mishra, A. Wei Yu, V. Zhao, Y. Huang, A. M. Dai, H. Yu, S. Petrov, Ed H. Chi, J. Dean, J. Devlin, A. Roberts, D. Zhou, Q. v Le, and J. Wei, (2022) Scaling Instruction-Finetuned Language Models. arXiv.

\noindent\hangindent=0.5cm [9] R. Umeike, N. Getty, F. Xia, and R. Stevens, (2025) Scaling Large Vision-Language Models for Enhanced Multimodal Comprehension in Biomedical Image Analysis. In \textit{2025 IEEE 22nd International Symposium on Biomedical Imaging (ISBI)}, pp. 1--4.

\noindent\hangindent=0.5cm [10] H. Liu, C. Li, Q. Wu, and Y. J. Lee, (2023) Visual Instruction Tuning. In \textit{Advances in Neural Information Processing Systems}, Curran Associates, Inc., pp. 34892--34916.

\noindent\hangindent=0.5cm [11] Q. Huang, H. Ren, P. Chen, G. Krzmanc, D. Zeng, P. Liang, and J. Leskovec, (2023) PRODIGY: enabling in-context learning over graphs. In \textit{Advances in Neural Information Processing Systems 36: Annual Conference on Neural Information Processing Systems 2023, NeurIPS 2023}, New Orleans, LA, USA, December 10--16.

\noindent\hangindent=0.5cm [12] Q. Dong, L. Li, D. Dai, C. Zheng, J. Ma, R. Li, H. Xia, J. Xu, Z. Wu, T. Liu, B. Chang, X. Sun, L. Li, and Z. Sui, (2023) A Survey on In-context Learning. arXiv.

\noindent\hangindent=0.5cm [13] T. B. Brown, B. Mann, N. Ryder, M. Subbiah, J. Kaplan, P. Dhariwal, A. Neelakantan, P. Shyam, G. Sastry, A. Askell, S. Agarwal, A. Herbert-Voss, G. Krueger, T. Henighan, R. Child, A. Ramesh, D. M. Ziegler, J. Wu, C. Winter, C. Hesse, M. Chen, E. Sigler, M. Litwin, S. Gray, B. Chess, J. Clark, C. Berner, S. McCandlish, A. Radford, I. Sutskever, and D. Amodei, (2020) Language Models are Few-Shot Learners. arXiv.

\noindent\hangindent=0.5cm [14] A. Grattafiori, A. Dubey, A. Jauhri, A. Pandey, A. Kadian, A. Al-Dahle, A. Letman, A. Mathur, A. Schelten, A. Vaughan
% ...
(2024) The Llama 3 Herd of Models. arXiv.

\noindent\hangindent=0.5cm [15] OpenAI, :, A. Hurst, A. Lerer, A. P. Goucher, A. Perelman, A. Ramesh, A. Clark, A. Ostrow, A. Welihinda, A. Hayes, A. Radford
% ...
(2024) GPT-4o System Card. arXiv.

\noindent\hangindent=0.5cm [16] O. Siméoni, H. V. Vo, M. Seitzer, F. Baldassarre, M. Oquab, C. Jose, V. Khalidov, M. Szafraniec, S. Yi, M. Ramamonjisoa, F. Massa, D. Haziza, L. Wehrstedt, J. Wang, T. Darcet, T. Moutakanni, L. Sentana, C. Roberts, A. Vedaldi, J. Tolan, J. Brandt, C. Couprie, J. Mairal, H. Jégou, P. Labatut, and P. Bojanowski, (2025) DINOv3. arXiv.

\noindent\hangindent=0.5cm [17] Z. Shao, P. Wang, Q. Zhu, R. Xu, J. Song, X. Bi, H. Zhang, M. Zhang, Y. K. Li, Y. Wu, and D. Guo, (2024) DeepSeekMath: Pushing the Limits of Mathematical Reasoning in Open Language Models. arXiv.

\noindent\hangindent=0.5cm [18] I. Beltagy, K. Lo, and A. Cohan, (2019) SciBERT: A Pretrained Language Model for Scientific Text. In \textit{Proceedings of the 2019 EMNLP-IJCNLP}, Hong Kong, China: Association for Computational Linguistics, Nov., pp. 3615--3620.

\noindent\hangindent=0.5cm [19] H. Wang, T. Fu, Y. Du, W. Gao, K. Huang, Z. Liu, P. Chandak, S. Liu, P. Van Katwyk, A. Deac, A. Anandkumar, K. Bergen, C. P. Gomes, S. Ho, P. Kohli, J. Lasenby, J. Leskovec, T.-Y. Liu, A. Manrai, D. Marks, B. Ramsundar, L. Song, J. Sun, J. Tang, P. Veličković, M. Welling, L. Zhang, C. W. Coley, Y. Bengio, and M. Zitnik, (2023) Scientific discovery in the age of artificial intelligence. \textit{Nature} \textbf{620}(7972):47--60.

\noindent\hangindent=0.5cm [20] P. Lewis, E. Perez, A. Piktus, F. Petroni, V. Karpukhin, N. Goyal, H. Küttler, M. Lewis, W. Yih, T. Rocktäschel, S. Riedel, and D. Kiela, (2020) Retrieval-Augmented Generation for Knowledge-Intensive NLP Tasks. arXiv.

\noindent\hangindent=0.5cm [21] E. J. Hu, Y. Shen, P. Wallis, Z. Allen-Zhu, Y. Li, S. Wang, L. Wang, and W. Chen, (2021) LoRA: Low-Rank Adaptation of Large Language Models. arXiv.

\noindent\hangindent=0.5cm [22] Y. Gao, Y. Xiong, X. Gao, K. Jia, J. Pan, Y. Bi, Y. Dai, J. Sun, M. Wang, and H. Wang, (2023) Retrieval-Augmented Generation for Large Language Models: A Survey. arXiv.

\noindent\hangindent=0.5cm [23] Daniel Han, Michael Han, and Unsloth team, (2023) Unsloth.

\noindent\hangindent=0.5cm [24] I. Loshchilov and F. Hutter, (2019) Decoupled Weight Decay Regularization. In \textit{7th International Conference on Learning Representations, ICLR 2019}, New Orleans, LA, USA, May 6--9.

%%%%%%%%%%%%%%%%%%%%%%%%%%%%%%%%%%%%%%%%%%%%%%%%%%%%%%%%%%%%

\appendix
\newpage
\section{Technical Appendices and Supplementary Material}

\subsection{Related Works}

Our work is situated at the intersection of foundation models for scientific discovery [18, 19] and domain adaptation techniques. While many studies explore either parameter-efficient fine-tuning (PEFT) techniques like LoRA [9] or in-context learning (ICL) with retrieval augmentation [20], a systematic comparison of these strategies in specialized, low-data scientific domains has been lacking.

\subsection{The PtychoBench Dataset}
This section provides a detailed description of the PtychoBench benchmark, which was curated to facilitate a reproducible and quantitative study of AI agents for ptychographic analysis.

\subsubsection{Data Curation and Partitioning} 
The dataset was constructed from an initial pool of 394 ptychographic reconstructions of experimental data acquired at the Advanced Photon Source, each annotated by a domain expert. After cleaning, the primary dataset consists of 391 samples. This full dataset was first partitioned into a training set of 312 samples and a test set of 79 samples. This primary 80/20 partition serves as the basis for the Artifact Detection task. Subsequently, the dataset for the Parameter Recommendation task was derived by creating subsets from these existing partitions. We filtered both the training and test sets to include only those instances which contained an expert's free-text recommendation label (caption\_param). This resulted in a final training set of 91 samples and a test set of 44 samples for the recommendation task. See Table 3 for sample data field.

\begin{table}[h!]
\centering
\scriptsize % Reduces the font size of the table
\caption{\textbf{Dataset Field Descriptions and Examples}: The table presents the structure and annotation schema of the PtychoBench dataset, detailing each field's purpose and usage in the VLM and LLM training and evaluation tasks.}
\label{tab:dataset_fields}
\setlength{\arrayrulewidth}{0.5pt} % Sets the width of all horizontal and vertical rules
\setlength{\tabcolsep}{4pt} % Adds horizontal space between columns
\renewcommand{\arraystretch}{1.5} % Adds vertical space between rows
\begin{tabular}{|>{\raggedright\arraybackslash}m{2cm}|>{\centering\arraybackslash}m{1.2cm}|>{\centering\arraybackslash}m{3.5cm}|>{\centering\arraybackslash}m{4cm}|>{\centering\arraybackslash}m{1.5cm}|}
\hline
\textbf{Field Name} & \textbf{Data Type} & \textbf{Description} & \textbf{Example Value} & \textbf{Usage} \\
\hline\hline
\verb|id| & Integer & Unique sample identifier & 2 & Both tasks \\
\hline
\verb|captioning| & String & Path to ptychographic reconstruction image & \texttt{/data/upload/1/af707e7b-...png} & VLM Training \\
\hline
\verb|beam_choice| & Categorical & Type of radiation beam used & X-ray & LLM Training \\
\hline
\verb|instrument_| \verb|choice| & Categorical & Specific beamline/instrument identifier & APS-2IDE-XFM & LLM Training \\
\hline
\verb|sample_text| & Categorical & Material/sample type being imaged & Integrated circuit & \textcolor{blue}{SSFS retrieval}, LLM Training\\
\hline
\verb|package_| \verb|choice| & Categorical & Reconstruction software package & pty-chi & LLM Training \\
\hline
\verb|artifact_| \verb|choice| & List & Expert-annotated visual artifacts present & [\textit{``Local distortion''}] & \textbf{VLM Target} \\
\hline
\verb|caption_obj| & String & Expert description of visual artifacts/quality & \textit{``Lines look zig-zag with discontinuities...''} & LLM Training \\
\hline
\verb|caption_| \verb|param| & String & Expert parameter recommendations & \textit{``Increase iterations or use smaller patterns...''} & \textbf{LLM Target} \\
\hline
\verb|recon_path_| \verb|text| & String & Full file path to reconstruction data & \texttt{/mnt/micdata1/2ide/2025-1/...} & Metadata \\
\hline
\verb|annotator| & Integer & Expert annotator identifier & 1 & Quality control \\
\hline
\verb|annotation_| \verb|id| & Integer & Annotation session identifier & 3 & Quality control \\
\hline
\verb|created_at| & Timestamp & Initial annotation timestamp & 2025-05-14T20:10:35Z & Metadata \\
\hline
\verb|updated_at| & Timestamp & Last modification timestamp & 2025-06-02T19:22:01Z & Metadata \\
\hline
\verb|lead_time| & Float & Annotation time in seconds & 180.11 & Quality metrics \\
\hline
\end{tabular}
\end{table}

\subsubsection{Task Definitions} 
The PtychoBench benchmark is designed for two sequential tasks central to autonomous characterization:
\begin{itemize}
\item \textbf{Artifact Detection}: A multi-modal, multi-label classification task where a Vision-Language Model (VLM) takes a ptychographic image as input and identifies a set of visual artifacts or defects.
\item \textbf{Parameter Recommendation}: A text-only, multi-label classification task where a Language Model (LLM) takes a textual description of observed artifacts and experimental conditions as input and recommends corrective actions.
\end{itemize}

\begin{figure}[h]
    \centering
    % First Row
    \begin{subfigure}{0.48\textwidth}
        \includegraphics[width=\textwidth]{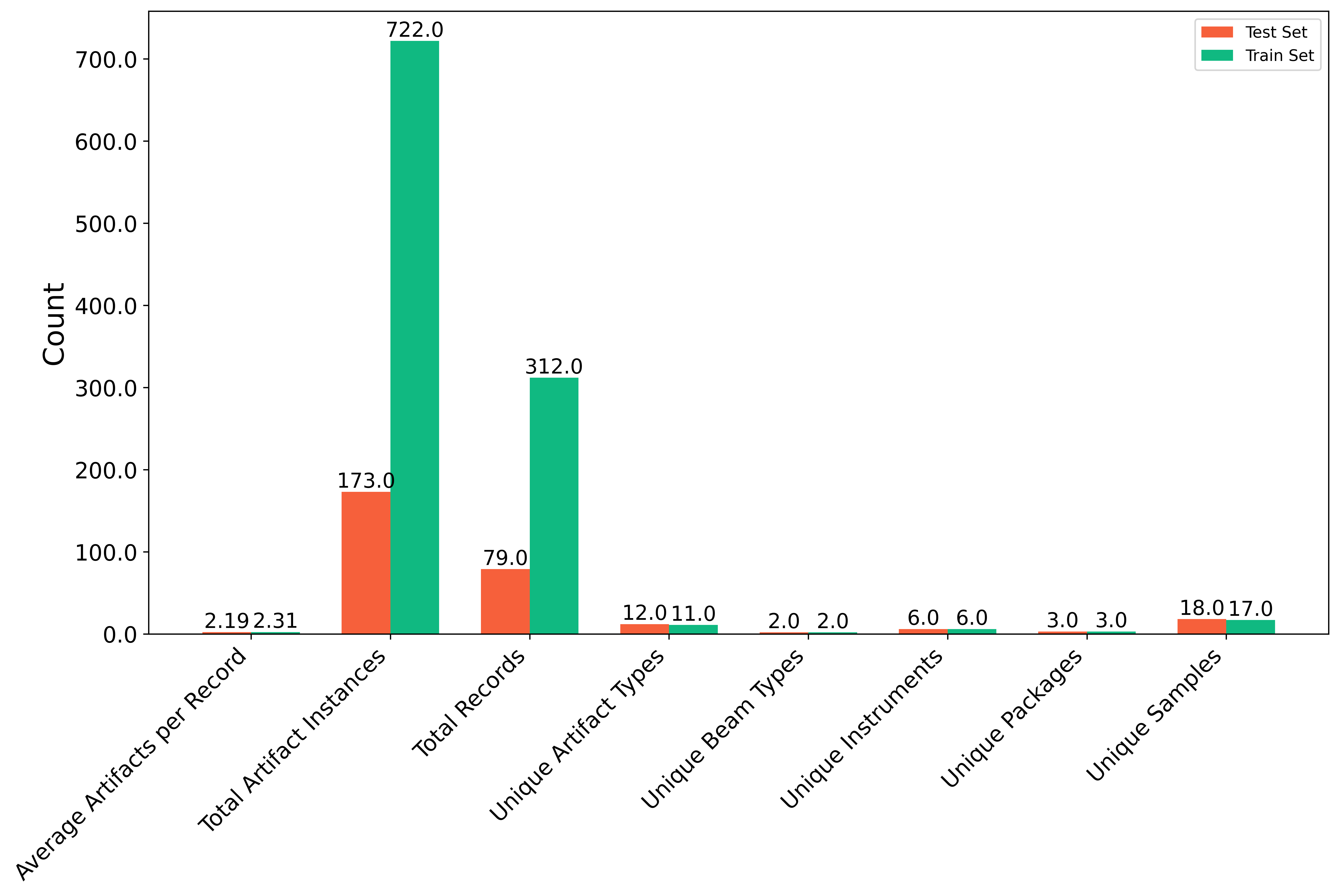}
        \caption{Key Summary Metrics}
        \label{fig:key_metrics}
    \end{subfigure}
    \hfill % This creates a horizontal space between the two images
    \begin{subfigure}{0.48\textwidth}
        \includegraphics[width=\textwidth]{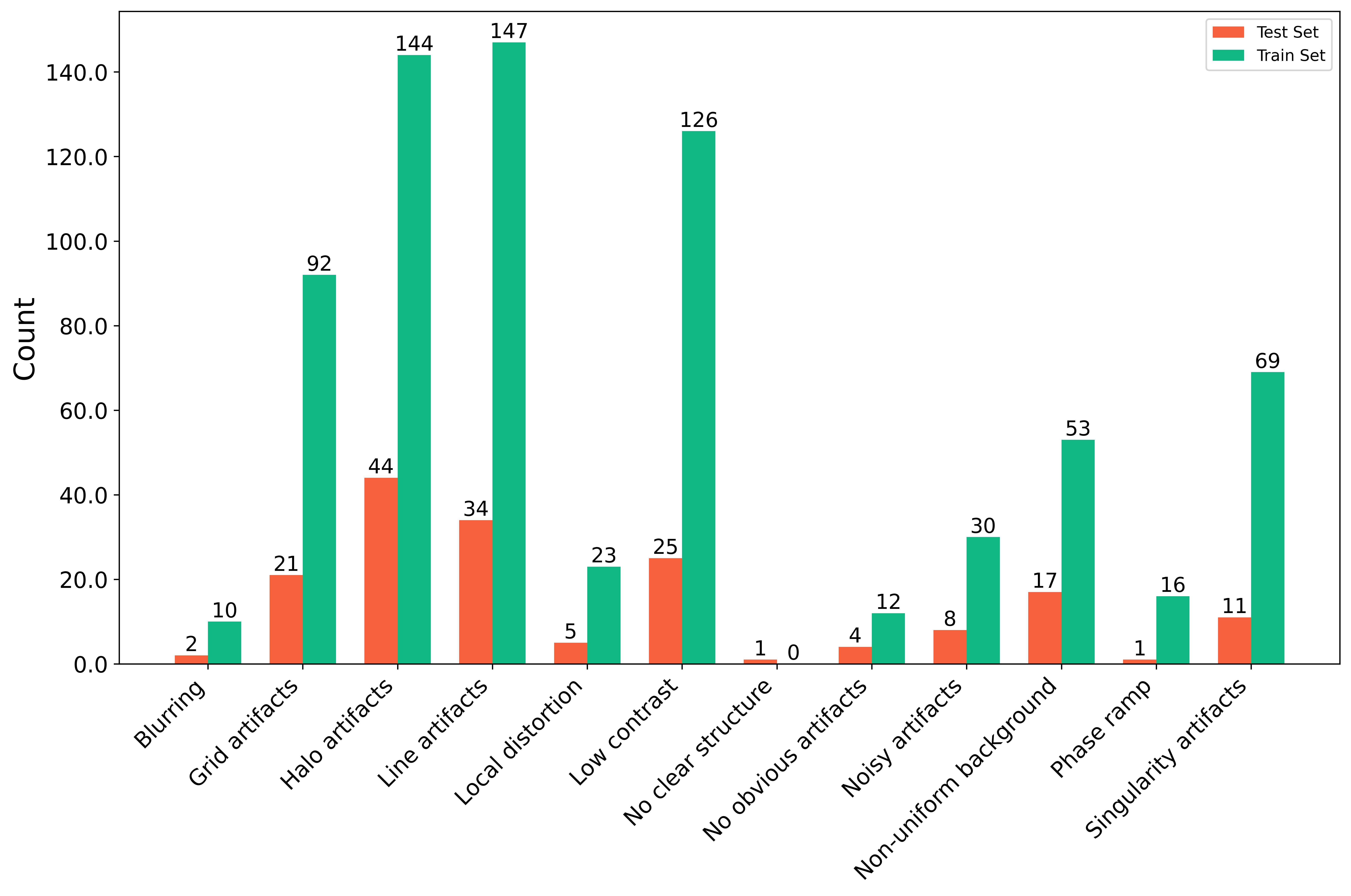}
        \caption{Artifact Type Distribution}
        \label{fig:artifact_counts}
    \end{subfigure}

    \vspace{0.5cm} % Optional vertical space between rows

    % Second Row
    \begin{subfigure}{0.48\textwidth}
        \includegraphics[width=\textwidth]{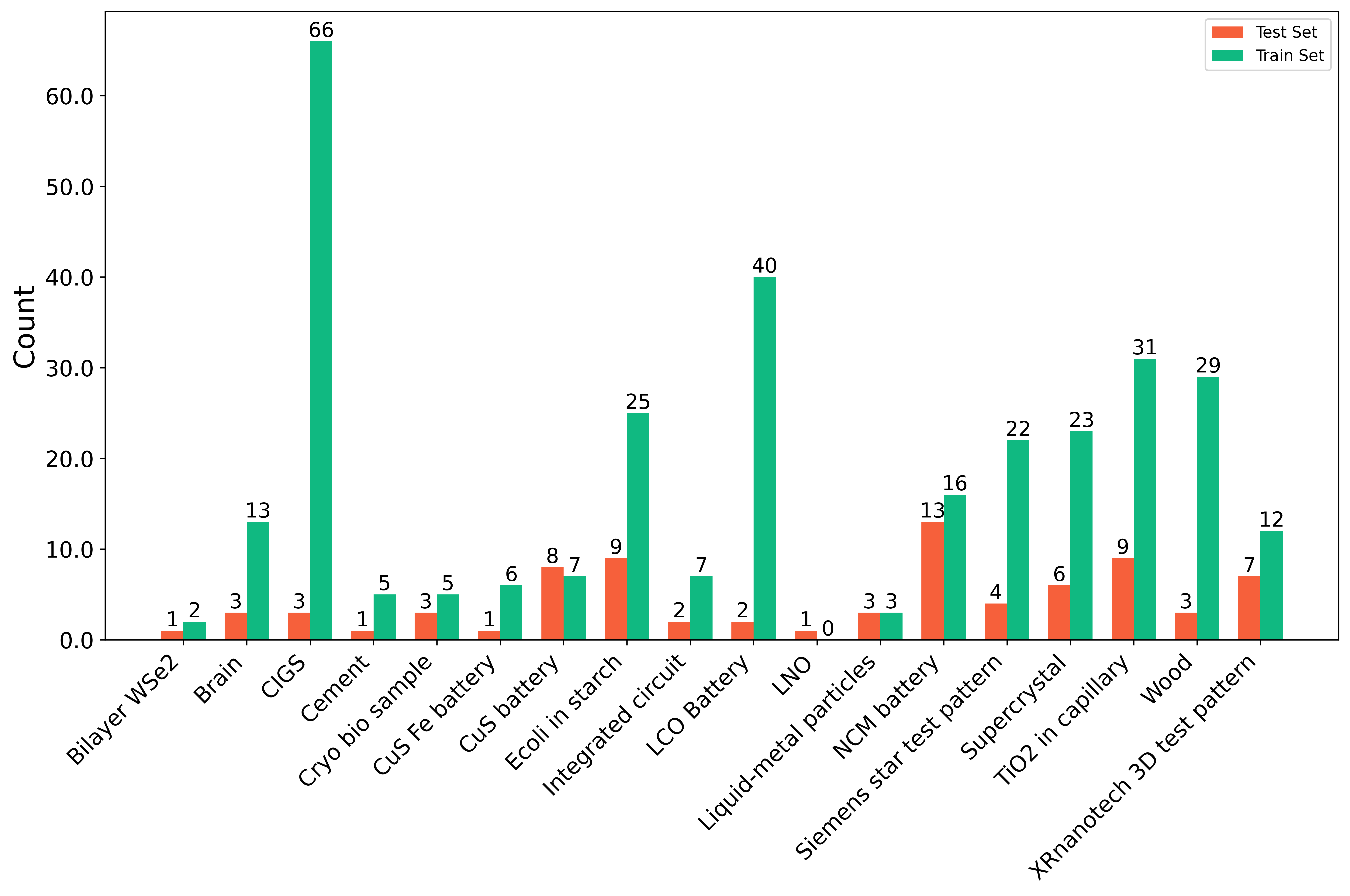}
        \caption{Sample Type Distribution}
        \label{fig:sample_counts}
    \end{subfigure}
    \hfill % This creates a horizontal space between the two images
    \begin{subfigure}{0.48\textwidth}
        \includegraphics[width=\textwidth]{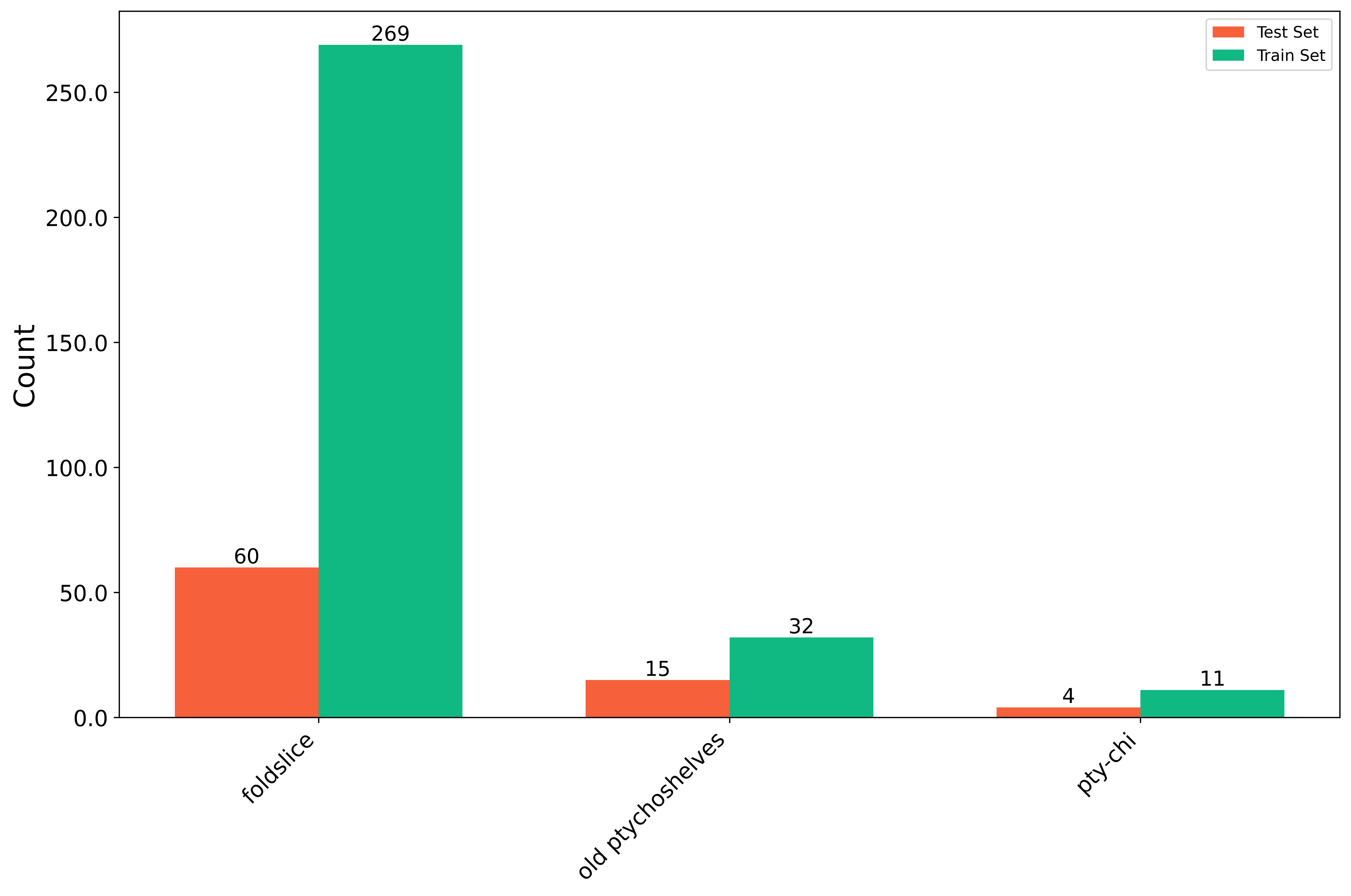}
        \caption{Software Package Distribution}
        \label{fig:package_counts}
    \end{subfigure}
    
    \caption{Statistical distributions of the PtychoBench dataset across the training and test sets. The summary plot (a) is followed by detailed breakdowns for artifact types (b), sample types (c), and software packages (d).}
    \label{fig:dataset_distributions}
\end{figure}

\subsubsection{Dataset Composition and Characteristics} 
The PtychoBench is characterized by its diversity and complexity, reflecting real-world experimental conditions. The full dataset spans 18 distinct sample types and 6 unique instruments. The Artifact Detection task presents a challenging multi-label problem. The 79 samples in the test set contain a total of 173 annotated artifact instances, averaging 2.19 artifacts per image. Across the entire dataset, the most prevalent of the 12 possible artifact types are Halo artifacts and Line artifacts. The Parameter Recommendation task is similarly complex, drawing from a rich context of metadata and observed issues. A complete statistical breakdown of the dataset composition, including the distributions for all categorical metadata as provided, is detailed in Figure 1 and Table 3.

\begin{table}[h!]
\centering
\scriptsize
\caption{\textbf{Task-Specific Data Usage}. The table details how different fields from the PtychoBench dataset are utilized in the VLM artifact detection task and LLM parameter recommendation task.}
\label{tab:task_specific_usage}
\setlength{\arrayrulewidth}{0.5pt}
\setlength{\tabcolsep}{3pt}
\renewcommand{\arraystretch}{2.0}
\begin{tabular}{|>{\raggedright\arraybackslash}m{2.2cm}|>{\centering\arraybackslash}m{3.9cm}|>{\centering\arraybackslash}m{2.5cm}|>{\centering\arraybackslash}m{2.3cm}|>{\centering\arraybackslash}m{1.5cm}|}
\hline
\textbf{Task} & \textbf{Input Features} & \textbf{Target Labels} & \textbf{Context Selection} & \textbf{Sample Size} \\
\hline\hline
\textbf{VLM Artifact Detection} &
captioning (image) + caption\_obj (description) [sample\_text, instrument\_choice,
beam\_choice] &
\textbf{artifact\_choice} (multi-label) &
sample\_text matching &
\textbf{391} \\
\hline\hline
\textbf{LLM Parameter Recommendation} &
prompt + context [beam\_choice, instrument\_choice, sample\_text, package\_choice, caption\_obj (artifact description)] &
\textbf{caption\_param} (text gen) &
sample\_text matching &
\textbf{135} \\
\hline
\end{tabular}
\end{table}

\subsection{Task Formulation and Prompt Engineering}
The prompts for both tasks were structured to emulate a real-world interaction with an expert AI agent, providing clear context and instructions. The data was formatted into a conversational structure with \textsc{user} and \textsc{assistant} roles.

\subsubsection{Task 1: LLM-based Parameter Recommendation}
This task is a text-only, multi-label classification problem simulating the decision-making step that follows artifact diagnosis.

\textbf{Prompt Template}: The prompt provides a rich textual context constructed by populating the template below with a sample's specific metadata.

\begin{tcolorbox}[title=Parameter Recommendation Prompt,
                  colback=red!5!white,
                  colframe=purple!75!black]
        *SAMPLE INFORMATION*:\newline

        Package Choice: {package}\newline
        Instrument Used: {instrument}\newline
        Sample Type: {sample}\newline
        Beam Type: {beam}\newline
        Expert Description: {caption\_obj}\newline
        Observed Artifacts: {artifacts}\newline
        Current Reconstruction Parameters:\newline
        {param\_info}\newline
        
        Based on this sample information, what specific parameter adjustments are needed to improve image quality?\newline
        
        *POSSIBLE PARAMETER UPDATES*:\newline
        A) Change number of batches to 1\newline
        B) Disable momentum acceleration\newline
        C) Disable multimodal update\newline
        D) Disable position correction\newline
        E) Enable momentum acceleration\newline
        F) Enable multimodal update\newline
        G) Enable position correction\newline
        H) Increase batch size\newline
        I) Increase number of OPR modes\newline
        J) Increase number of probe modes\newline
        K) Increase the number of iterations\newline
        L) No changes needed\newline
        M) Recenter diffraction patterns\newline
        N) Reduce batch size\newline
        O) Reduce diffraction pattern size by factor of 2\newline
        P) Reduce or disable regularization\newline
        Q) Try other diffraction pattern orientations\newline
        R) Turn off affine constraint\newline
        S) Turn off variable probe correction (set OPR modes to 0)\newline
        T) Turn on variable probe correction (set OPR modes to 1)\newline
        U) Use Gaussian noise model\newline
        V) Use compact batch selection scheme\newline
        W) Use multislice model\newline
        X) Use sparse batch selection scheme\newline
        
        Your task:\newline
        1. Select all correct options (A, B, C, ...) and list them as a comma-separated list inside <answer></answer> tags (e.g., <answer>E, J, V</answer>).\newline
        2. The <answer></answer> tags must contain ONLY the letters. Do not include any explanations or other text.\newline

\end{tcolorbox}
\begin{tcolorbox}[colframe=blue!75!black, colback=white, boxrule=1pt]
         \textbf{Ground-Truth Generation:} The ground-truth labels for the assistant response were generated via a complex procedure. The expert's original free-text advice (the caption\_param field) was programmatically mapped to the corresponding alphabetic choices (A-X) using a curated set of regular expression patterns. This allowed us to convert natural language instructions (e.g., "Increase the number of iterations to 2000") into a standardized, multi-label format (e.g., <answer>K</answer>).
\end{tcolorbox}

\subsubsection{Task 2: VLM-based Artifact Detection}
This task is formulated as a multi-modal, multi-label classification problem where a VLM identifies visual artifacts in a ptychographic image.

\textbf{Prompt Template}: The user prompt consists of an image paired with the following text instruction:

\begin{tcolorbox}[title=Artifact Detection Prompt,
                  colback=red!5!white,
                  colframe=purple!75!black]
        You are an expert in X-ray ptychography image analysis. What artifacts are visible?\newline 
        
        Possible artifacts include: Grid artifacts, Halo artifacts, Line artifacts, Local distortion, Low contrast, No clear structure, Noisy artifacts, Non-uniform background, Phase ramp, Singularity artifacts, Blurring.\newline
        
        Format your response as: <answer>[Comma-separated list of artifacts, or "No obvious artifacts" if the image quality is good]</answer> Provide your answers now:
\end{tcolorbox}
\begin{tcolorbox}[colframe=blue!75!black, colback=white, boxrule=1pt]
        \textbf{Ground-Truth Generation:} The assistant response in the training data contains the expert-annotated list of artifacts (from the artifact\_choice field), formatted within <answer> tags as requested by the prompt.
\end{tcolorbox}

\subsection{Model and Training Details}
This section provides the specific identifiers, libraries, and hyperparameters used for all experiments.

\subsubsection{Models}
All open-weight models were used in their full-precision versions. The specific Hugging Face identifiers used are:
\begin{itemize}
\item \textbf{VLMs:} meta-llama/Llama-3.2-11B-vision and meta-llama/Llama-3.2-90B-vision.
\item \textbf{LLMs:} meta-llama/Meta-Llama-3.1-8B-Instruct and meta-llama/Meta-Llama-3.1-70B-Instruct.
\item \textbf{Library}: Both the LLM and VLM training processes utilized the Unsloth library [23]. 
\item \textbf{Proprietary Baseline:} OpenAI's \texttt{gpt-4o} model was accessed via their API.
\item \textbf{Vision Baseline:} The Meta AI \texttt{DINOv3} model was used as a fixed feature extractor for the traditional computer vision baseline.
\end{itemize}

% Required packages: \usepackage{booktabs}

\begin{table}[h!]
\centering
\small
\caption{\textbf{SFT Hyperparameters for Vision-Language Models (VLMs)}. Training configuration used for supervised fine-tuning of Llama 3.2-Vision models on the artifact detection task.}
\label{tab:vlm_hyperparams}
\setlength{\tabcolsep}{12pt}
\renewcommand{\arraystretch}{1.3}
\begin{tabular}{>{\raggedright\arraybackslash}p{4.5cm} >{\raggedleft\arraybackslash}p{3cm}}
\toprule
\textbf{Hyperparameter} & \textbf{Value} \\
\midrule
SFT Epoch (11B model) & 50 \\
SFT Epoch (90B model) & ~20 \\
\midrule
\texttt{learning\_rate} & 2e-4 \\
\texttt{lora\_r} & 16 \\
\texttt{lora\_alpha} & 16 \\
\texttt{lora\_dropout} & 0 \\
\texttt{bias} & "none" \\
\texttt{optim} & "adamw\_8bit" \\
\texttt{weight\_decay} & 0.01 \\
\midrule
\texttt{max\_seq\_length} & 2048 \\
\texttt{per\_device\_train\_batch\_size} & 1 \\
\texttt{gradient\_accumulation\_steps} & 8 \\
\bottomrule
\end{tabular}
\end{table}

\begin{table}[h!]
\centering
\small
\caption{\textbf{SFT Hyperparameters for Language Models (LLMs)}. Training configuration used for supervised fine-tuning of Llama 3.1 models on the parameter recommendation task.}
\label{tab:llm_hyperparams}
\setlength{\tabcolsep}{12pt}
\renewcommand{\arraystretch}{1.3}
\begin{tabular}{>{\raggedright\arraybackslash}p{4.5cm} >{\raggedleft\arraybackslash}p{4.5cm}}
\toprule
\textbf{Hyperparameter} & \textbf{Value} \\
\midrule
\texttt{num\_epochs} & 50 \\
\texttt{learning\_rate} & 2e-4 \\
\midrule
\texttt{lora\_r} & 16 \\
\texttt{lora\_alpha} & 16 \\
\texttt{target\_modules} & \texttt{["q\_proj", "k\_proj", "v\_proj",} \\
& \texttt{"o\_proj", "gate\_proj", "up\_proj",} \\
& \texttt{"down\_proj"]} \\
\texttt{lora\_dropout} & 0 \\
\texttt{bias} & "none" \\
\midrule
\texttt{optim} & "adamw\_8bit" \\
\texttt{weight\_decay} & 0.01 \\
\midrule
\texttt{max\_seq\_length} & 2048 \\
\texttt{per\_device\_train\_batch\_size} & 4 \\
\texttt{gradient\_accumulation\_steps} & 8 \\
\bottomrule
\end{tabular}
\end{table}

\subsubsection{Supervised Fine-Tuning (SFT) Protocol}
We employed parameter-efficient fine-tuning (PEFT) using Low-Rank Adaptation (LoRA). The loss function for our multi-label tasks was Binary Cross-Entropy with Logits Loss. The detailed hyperparameters are provided in Table \ref{tab:vlm_hyperparams} and Table \ref{tab:llm_hyperparams}.

\subsubsection{In-Context Learning (ICL) Protocol}
For the Sample-Specific Few-Shot (SSFS) strategy, examples were selected from the training set based on an exact match of the sample type metadata field with the test sample. If fewer than k examples with a matching sample type were available, the remaining slots in the prompt were filled by randomly sampling from the rest of the training set, ensuring the test sample itself was excluded.

\newpage
\subsection{Results}

\subsubsection{Comprehensive Results with Statistical Validation}
The main paper presents the mean and standard deviation for our experimental results in Tables \ref{table:artifact_detection_mean_std} and \ref{table:param_reco_mean_std}. To provide a complete picture of statistical significance and address the variability from a limited test set, this section provides the full 95\% confidence intervals (CIs) for all VLM and LLM experiments. These CIs were computed via bootstrapping with 10,000 resamples. The tables allow for a direct comparison of the performance ranges between different models and strategies, directly supporting the statistical conclusions drawn in the main text.

\begin{table}[h!]
\centering
\scriptsize % Reduces the font size of the table
\caption{VLM Performance on Artifact Detection (95\% Confidence Interval)} 
\label{table:artifact_detection_ci}
\setlength{\arrayrulewidth}{0.7pt} % Sets the width of all horizontal and vertical rules
\setlength{\tabcolsep}{5.7pt} % Adds horizontal space between columns
\renewcommand{\arraystretch}{1.8} % Adds vertical space between rows
\begin{tabular}{|l|c|c|c|c|c|c|}
\hline
\textbf{Model} & \textbf{Fewshot Strategy} & \textbf{0-shot\protect\footnotemark} & \textbf{1-shot} & \textbf{3-shot} & \textbf{5-shot} & \textbf{7-shot} \\
\hline
\multirow{2}{*}{\textbf{SFT Llama 3.2-Vision 11B}} & RFS & \multirow{2}{*}{[0.412 - 0.573]} & [0.251 - 0.392] \textcolor{red}{$\downarrow$} & [0.239 - 0.399] & [0.308 - 0.468] & [0.315 - 0.475] \\
\cline{2-2}\cline{4-7}
& SSFS & & \textbf{[0.458 - 0.631]} & \textbf{[0.547 - 0.720]} & \textbf{[0.581 - 0.748]} & \textbf{[0.636 - 0.784]} \\
\hline\hline
\multirow{2}{*}{\textbf{Base Llama 3.2-Vision 11B}} & RFS & \multirow{2}{*}{[0.168 - 0.266]} & [0.188 - 0.320] & [0.183 - 0.321] & [0.248 - 0.377] & [0.270 - 0.412] \\
\cline{2-2}\cline{4-7}
& SSFS & & \textbf{[0.512 - 0.696]} & \textbf{[0.535 - 0.715]} & \textbf{[0.610 - 0.773]} & \textbf{[0.615 - 0.769]} \\
\hline\hline
\multirow{2}{*}{\textbf{SFT Llama 3.2-Vision 90B}} & RFS & \multirow{2}{*}{[0.348 - 0.509]} & [0.264 - 0.403] \textcolor{red}{$\downarrow$} & [0.337 - 0.479] & [0.326 - 0.473] & [0.377 - 0.516] \\
\cline{2-2}\cline{4-7}
& SSFS & & \textbf{[0.500 - 0.668]} & \textbf{[0.504 - 0.683]} & \textbf{[0.589 - 0.749]} & \textbf{[0.655 - 0.796]} \\
\hline\hline
\multirow{2}{*}{\textbf{Base Llama 3.2-Vision 90B}} & RFS & \multirow{2}{*}{[0.000 - 0.050]} & [0.189 - 0.322] & [0.109 - 0.248] & [0.040 - 0.147] & [0.083 - 0.229] \\
\cline{2-2}\cline{4-7}
& SSFS & & \textbf{[0.514 - 0.700]} & \textbf{[0.525 - 0.723]} & \textbf{[0.525 - 0.713]} & \textbf{[0.626 - 0.779]} \\
\hline\hline
\multirow{2}{*}{\textbf{\textcolor{blue}{GPT-4o}}} & RFS & \multirow{2}{*}{[0.126 - 0.226]} & [0.117 - 0.209] & [0.137 - 0.242]  & [0.189 - 0.308] & [0.206 - 0.357] \\
\cline{2-2}\cline{4-7}
& SSFS & & \textbf{[0.262 - 0.410]} & \textbf{[0.441 - 0.614]} & \textbf{[0.546 - 0.720]} & \textbf{[0.571 - 0.739]} \\
% \hline\hline
% \textbf{\textcolor{purple}{DINOv3}} & - &  -- & - & - & - & - \\
\hline
\end{tabular}
\end{table}

\footnotetext{0-shot CIs from SSFS and RFS bootstrap runs differed negligibly; for simplicity, a single representative interval is reported.}

\begin{table}[h!]
\centering
\scriptsize % Reduces the font size of the table
\caption{LLM Performance on Parameter Recommendation (95\% Confidence Intervals)}
\label{table:param_reco_ci_only}
\setlength{\arrayrulewidth}{0.7pt} % Sets the width of all horizontal and vertical rules
\setlength{\tabcolsep}{7.2pt} % Adds horizontal space between columns
\renewcommand{\arraystretch}{1.8} % Adds vertical space between rows
\begin{tabular}{|l|c|c|c|c|c|c|}
\hline
\textbf{Model} & \textbf{Fewshot Strategy} & \textbf{0-shot} & \textbf{1-shot} & \textbf{3-shot} & \textbf{5-shot} & \textbf{7-shot} \\
\hline
\multirow{2}{*}{\textbf{SFT Llama 3.1 8B}} & RFS & \multirow{2}{*}{[0.370 - 0.572]} & [0.247 - 0.389] \textcolor{red}{$\downarrow$} & [0.328 - 0.502] & [0.394 - 0.598] & [0.477 - 0.630] \\
\cline{2-2}\cline{4-7}
& SSFS & & \textbf{[0.225 - 0.417]} & \textbf{[0.479 - 0.722]} & \textbf{[0.549 - 0.770]} & \textbf{[0.587 - 0.797]} \\
\hline\hline
\multirow{2}{*}{\textbf{Base Llama 3.1 8B}} & RFS & \multirow{2}{*}{[0.051 - 0.136]} & [0.114 - 0.258] & [0.205 - 0.337] & [0.359 - 0.546] & [0.297 - 0.481] \\
\cline{2-2}\cline{4-7}
& SSFS & & \textbf{[0.337 - 0.483]} & \textbf{[0.491 - 0.715]} & \textbf{[0.533 - 0.792]} & \textbf{[0.593 - 0.814]} \\
\hline\hline
\multirow{2}{*}{\textbf{SFT Llama 3.1 70B}} & RFS & \multirow{2}{*}{[0.734 - 0.928]} & [0.364 - 0.545] \textcolor{red}{$\downarrow$} & [0.411 - 0.573] & [0.427 - 0.621] & [0.496 - 0.689] \\
\cline{2-2}\cline{4-7}
& SSFS & & \textbf{[0.437 - 0.599]}  & \textbf{[0.508 - 0.703]} & \textbf{[0.593 - 0.764]} & \textbf{[0.597 - 0.781]} \\
\hline\hline
\multirow{2}{*}{\textbf{Base Llama 3.1 70B}} & RFS & \multirow{2}{*}{[0.189 - 0.266]} & [0.193 - 0.343] & [0.273 - 0.452] & [0.436 - 0.652] & [0.463 - 0.684] \\
\cline{2-2}\cline{4-7}
& SSFS & & \textbf{[0.446 - 0.634]} & \textbf{[0.664 - 0.820]} & \textbf{[0.697 - 0.861]} & \textbf{[0.755 - 0.923]} \\
\hline\hline
\multirow{2}{*}{\textbf{\textcolor{blue}{GPT-4o}}} & RFS & \multirow{2}{*}{[0.255 - 0.373]} & [0.269 - 0.418] & [0.240 - 0.415]  & [0.423 - 0.613] & [0.456 - 0.650] \\
\cline{2-2}\cline{4-7}
& SSFS & & \textbf{[0.521 - 0.709]} & \textbf{[0.614 - 0.796]} & \textbf{[0.671 - 0.851]} & \textbf{[0.701 - 0.854]} \\
\hline
\end{tabular}
\end{table}

\newpage
\subsubsection{Per-Class F1-Score for Artifact Detection (SFT-90B + SSFS)}
To provide a more granular analysis of model performance, especially given the inherent class imbalance of the dataset, we provide a per-class F1-score analysis for our best-performing visual model (SFT-90B with SSFS) in Table \ref{tab:per_class_f1}. The results show that the model's performance generally improves with more in-context examples ($k$) and is strongest on the most well-represented classes (e.g., "Halo artifacts," "Low contrast"). The model struggles with extremely rare classes (e.g., "No clear structure," "Phase ramp"), which have only one test sample. This imbalance is the primary driver for the gap between the Micro-F1 and Macro-F1 scores reported in the main paper.

\begin{table}[h!]
\centering
\scriptsize % Reduces the font size of the table
\caption{Per-Class F1-Scores for SFT-90B (SSFS) on Artifact Detection. N indicates the number of test samples (out of 79) that contain the artifact.}
\label{tab:per_class_f1}
\setlength{\arrayrulewidth}{0.7pt} % Sets the width of all horizontal and vertical rules
\setlength{\tabcolsep}{11.2pt} % Adds horizontal space between columns
\renewcommand{\arraystretch}{1.2} % Adds vertical space between rows
\begin{tabular}{|l|r|c|c|c|c|c|}
\hline
\textbf{Artifact Class} & \textbf{N} & \textbf{0-shot F1} & \textbf{1-shot F1} & \textbf{3-shot F1} & \textbf{5-shot F1} & \textbf{7-shot F1} \\
\hline\hline
Halo artifacts & 44 & 0.47 & 0.72 & 0.72 & 0.81 & 0.90 \\
\hline
Line artifacts & 34 & 0.53 & 0.64 & 0.64 & 0.60 & 0.67 \\
\hline
Low contrast & 25 & 0.52 & 0.68 & 0.67 & 0.84 & 0.88 \\
\hline
Grid artifacts & 21 & 0.40 & 0.48 & 0.50 & 0.56 & 0.67 \\
\hline
Non-uniform background & 17 & 0.49 & 0.64 & 0.60 & 0.72 & 0.75 \\
\hline
Singularity artifacts & 11 & 0.40 & 0.58 & 0.62 & 0.59 & 0.64 \\
\hline
Noisy artifacts & 8 & 0.29 & 0.38 & 0.46 & 0.80 & 0.63 \\
\hline
Local distortion & 5 & 0.00 & 0.00 & 0.00 & 0.22 & 0.00 \\
\hline
No obvious artifacts & 4 & 0.00 & 0.00 & 0.00 & 0.00 & 0.00 \\
\hline
Blurring & 2 & 0.00 & 0.00 & 0.00 & 0.00 & 0.00 \\
\hline
No clear structure & 1 & 0.00 & 0.00 & 0.00 & 0.00 & 0.00 \\
\hline
Phase ramp & 1 & 0.00 & 0.00 & 0.00 & 0.00 & 0.00 \\
\hline\hline
\textbf{Overall (Micro F1)} & \textbf{173} & \textbf{0.43} & \textbf{0.59} & \textbf{0.60} & \textbf{0.67} & \textbf{0.73} \\
\hline
\end{tabular}
\end{table}

\newpage
\subsubsection{Per-Sample-Type Performance on Parameter Recommendation}

While the preceding analysis addresses class imbalance, this section provides a scientific breakdown of performance by material category (sample\_type). We analyzed the best-performing LLM (Base Llama 3.1-70B with SSFS) to understand which ptychographic samples are more challenging for the model. The performance, as shown in Table \ref{tab:per_sample_f1}, is highly dependent on the material being analyzed.

\begin{table}[h!]
\centering
\scriptsize % Reduces the font size of the table
\caption{Per-Sample-Type Micro-F1 Scores for Base Llama 3.1-70B with SSFS on the Parameter Recommendation task. N indicates the number of test samples for each type.}
\label{tab:per_sample_f1}
\setlength{\arrayrulewidth}{0.7pt} % Sets the width of all horizontal and vertical rules
\setlength{\tabcolsep}{11.2pt} % Adds horizontal space between columns
\renewcommand{\arraystretch}{1.2} % Adds vertical space between rows
\begin{tabular}{|l|r|c|c|c|c|c|}
\hline
\textbf{Sample Type} & \textbf{N} & \textbf{0-shot F1} & \textbf{1-shot F1} & \textbf{3-shot F1} & \textbf{5-shot F1} & \textbf{7-shot F1} \\
\hline\hline
Ecoli in starch & 9 & 0.301 & 0.771 & 0.966 & 0.941 & 0.978 \\
\hline
TiO2 in capillary & 9 & 0.233 & 0.654 & 0.800 & 0.783 & 0.973 \\
\hline
Siemens star test pattern & 4 & 0.320 & 0.480 & 0.889 & 0.960 & 0.923 \\
\hline
XRnanotech 3D test pattern & 7 & 0.213 & 0.278 & 0.462 & 0.571 & 0.684 \\
\hline
Supercrystal & 6 & 0.108 & 0.222 & 0.429 & 0.417 & 0.500 \\
\hline
Liquid-metal particles & 3 & 0.250 & 0.714 & 0.889 & 1.000 & 1.000 \\
\hline
Integrated circuit & 2 & 0.167 & 0.000 & 0.286 & 0.667 & 0.889 \\
\hline
NCM battery & 1 & 0.250 & 0.750 & 0.750 & 1.000 & 1.000 \\
\hline
Bilayer WSe2 & 1 & 0.200 & 0.500 & 0.333 & 0.000 & 0.000 \\
\hline
CuS battery & 1 & 0.000 & 0.000 & 0.500 & 0.667 & 0.500 \\
\hline
LNO & 1 & 0.000 & 0.000 & 0.000 & 0.000 & 0.000 \\
\hline\hline
\textbf{Overall (Micro F1)} & \textbf{44} & \textbf{0.229} & \textbf{0.544} & \textbf{0.747} & \textbf{0.785} & \textbf{0.848} \\
\hline
\end{tabular}
\end{table}

The results in Table \ref{tab:per_sample_f1} show a strong correlation between the number of available test samples (N) and model performance, particularly for sample types with very few instances. The model consistently achieves near-perfect scores on types with many examples (e.g., Ecoli in starch) but struggles on types with only a single test sample (e.g., LNO, CuS battery). This suggests that while ICL is highly effective, its performance is still sensitive to the diversity and representation of the underlying data distribution, which is a key area for future work in dataset augmentation and expansion.

%%%%%%%%%%%%%%%%%%%%%%%%%%%%%%%%%%%%%%%%%%%%%%%%%%%%%%%%%%%%

\end{document}